\documentclass[runningheads]{llncs}
\usepackage{graphicx}

\usepackage{tikz}
\usepackage{comment}

\usepackage{amsmath,amssymb} 
\usepackage{color}

\usepackage[accsupp]{axessibility}  

\usepackage[ruled]{algorithm2e}
\usepackage{multirow}
\newcommand{\tabincell}[2]{\begin{tabular}{@{}#1@{}}#2\end{tabular}} 
\usepackage{booktabs}
\usepackage{bm}

\begin{document}
\pagestyle{headings}
\mainmatter
\def\ECCVSubNumber{1580}  

\title{Patch Similarity Aware Data-Free Quantization for Vision Transformers} 


\titlerunning{Patch Similarity Aware Data-Free Quantization for Vision Transformers}

\author{Zhikai Li\inst{1,2}\and Liping Ma\inst{1}\and Mengjuan Chen\inst{1}\and Junrui Xiao\inst{1,2}\and Qingyi Gu\inst{1}\thanks{Corresponding author.}} 
\authorrunning{Z. Li et al.}
%
\institute{Institute of Automation, Chinese Academy of Sciences \and
School of Artificial Intelligence, University of Chinese Academy of Sciences \\
\email{\{lizhikai2020,liping.ma,chenmengjuan2016,xiaojunrui2020,qingyi.gu\}@ia.ac.cn}}
\maketitle

\begin{abstract}
Vision transformers have recently gained great success on various computer vision tasks; nevertheless, their high model complexity makes it challenging to deploy on resource-constrained devices. Quantization is an effective approach to reduce model complexity, and data-free quantization, which can address data privacy and security concerns during model deployment, has received widespread interest. Unfortunately, all existing methods, such as BN regularization, were designed for convolutional neural networks and cannot be applied to vision transformers with significantly different model architectures. In this paper, we propose PSAQ-ViT, a Patch Similarity Aware data-free Quantization framework for Vision Transformers, to enable the generation of ``realistic" samples based on the vision transformer's unique properties for calibrating the quantization parameters. Specifically, we analyze the self-attention module's properties and reveal a general difference (patch similarity) in its processing of Gaussian noise and real images. The above insights guide us to design a relative value metric to optimize the Gaussian noise to approximate the real images, which are then utilized to calibrate the quantization parameters. Extensive experiments and ablation studies are conducted on various benchmarks to validate the effectiveness of PSAQ-ViT, which can even outperform the real-data-driven methods. Code is available at: \url{https://github.com/zkkli/PSAQ-ViT}.

\end{abstract}

\section{Introduction}
With the great success on natural language processing applications, transformer-based models have also demonstrated superior performance on a variety of computer vision tasks \cite{khan2021transformers,han2020survey}. However, vision transformers typically employ complicated model architectures with extremely high memory footprints and computational overheads to accomplish the powerful representational capabilities, posing significant challenges for their deployment and real-time inference on resource-constrained edge devices \cite{tang2021patch,jia2021efficient,liu2021post}. Thus, the compression technique for vision transformers is highly desired for real-world applications.

Model quantization, which converts 32-bit floating-point parameters (weights and activations) to low-precision values, is regarded as a prevalent approach to reduce the complexity of neural networks and accelerate their inference phase \cite{krishnamoorthi2018quantizing,gholami2021survey}. To mitigate the accuracy degradation, almost all quantization methods require access to the original dataset for re-training/fine-tuning the model parameters \cite{zhang2018lq,elthakeb2020gradient,chin2020one,hubara2016binarized,rastegari2016xnor}. Unfortunately, in scenarios involving sensitive data ($e$.$g$., medical and bio-metric data), these methods are no longer applicable due to the unavailability of the original dataset \cite{xu2020generative,zhang2021diversifying}. Therefore, data-free quantization is regarded as a potential and practice scheme \cite{cai2020zeroq,zhong2021intraq}.

The main idea of data-free quantization is to generate samples that can match the real-data distribution based on the prior information of the pre-trained full-precision (FP) model, and then utilize these samples to calibrate the quantization parameters. The key issue is how to generate effective and meaningful samples to ensure the calibration accuracy. A notable line of research proposes batch normalization (BN) regularization \cite{cai2020zeroq,zhang2021diversifying}, which states that the statistics ($i$.$e$., the mean and standard deviation) encoded in the BN layers can represent the distribution of original training data. These methods, however, are only applicable to convolutional neural networks (CNNs) and not to vision transformers, because the latter employs layer normalization (LN), which does not store any previous information like BN. As a result, existing methods cannot be extended and migrated well due to significant differences in model architecture, leaving data-free quantization for vision transformers as a gap.

\begin{figure}
	\centering
	\includegraphics[width=0.95\linewidth]{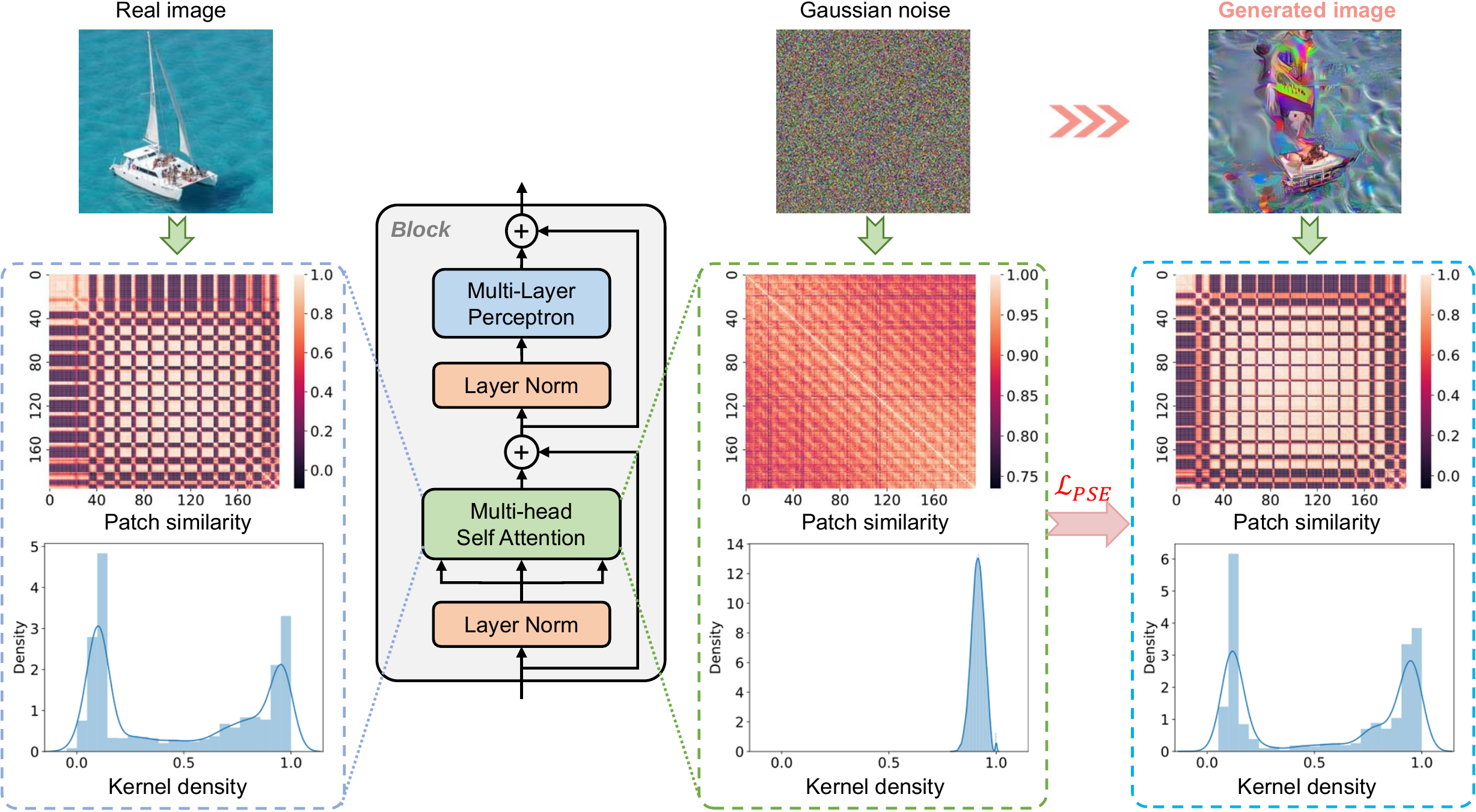}
	\caption{Illustration of the proposed sample generation approach. When the input is Gaussian noise, patches are grouped into one category (foreground or background), leading to homogeneous patch similarity and a unimodal kernel density curve. Our generated image can potentially represent the real-image features, producing diverse patch similarity and a bimodal kernel density curve, where the left and right peaks describe inter- and intra-category similarity, respectively.}
	\label{fig:1}
\end{figure}

In this paper, we are motivated to address the above issues, focusing on the following challenge: \emph{how to effectively generate ``realistic" samples based on the vision transformer's unique properties?} 
Since there is no elegant absolute value metric like BN statistics, we intend to investigate the general difference in model inference when the input is Gaussian noise and a real image, and then accordingly design a \emph{relative value} metric to optimize the noise. 
As stated in \cite{dosovitskiy2020image}, in the training phase, the self-attention module is designed to extract the important information from the training data, $i$.$e$., to identify the foreground from the background, so that the model can make a good decision.
Accordingly, in the inference phase of the pre-trained model, when the input is a real image, the foreground patches and background patches can produce different responses, thus the self-attention module has a diverse patch similarity ($i$.$e$. the similarity between the responses in the patch dimension); in contrast, the responses to Gaussian noise, whose foreground and background are hard to distinguish, are homogeneous, as shown in Fig. \ref{fig:1}.

With the above analysis, we propose PSAQ-ViT, a Patch Similarity Aware data-free Quantization framework for Vision Transformers. Specifically, we utilize the differential entropy of patch similarity to quantify the diversity of responses, which is calculated via kernel density estimation that can ensure gradient back-propagation. Then, the differential entropy is used as the objective function to optimize the Gaussian noise to approximate the real image. Finally, the generated samples are utilized to calibrate the parameters of the quantized vision transformers.

To be specific, our contributions are as follows:
\begin{itemize}
	\item From an in-depth analysis of the self-attention module, we reveal a general difference in its processing of Gaussian noise and real images, $i$.$e$., a substantially distinct diversity of patch similarity. This general difference demonstrates the intrinsic properties of vision transformers' image perception and provides some insights for sample generation.
	\item With the above insights, we propose PSAQ-ViT, in which we design a relative value metric to optimize the Gaussian image to reduce the general difference and thus approximate the real images, and then utilize them to calibrate the quantization parameters. To the best of our knowledge, this is the first work to quantify vision transformers without any real-world data.
	\item Extensive experiments on various benchmark models are conducted to demonstrate the effectiveness of PSAQ-ViT, which can generate ``realistic" samples and thus enable the outstanding performance of data-free quantization for vision transformers, even outperforming real-data-driven methods.
\end{itemize}

\section{Related Works}
\subsection{Vision transformers}
Vision transformers, which utilize global information based on self-attention modules, have recently achieved great success on various computer vision tasks \cite{arnab2021vivit,zhou2021deepvit,chen2021transformer,wu2021rethinking,han2021transformer}. ViT \cite{dosovitskiy2020image} is the first pure vision transformer model, which reshapes the image into a sequence of flattened 2D patches as the input, achieving better performance than CNNs on image classification tasks. Following ViT, DeiT \cite{touvron2021training} introduces a teacher-student procedure based on a distillation token, which can achieve competitive results on ImageNet with no external data. Swin Transformer \cite{liu2021swin} presents a hierarchical design with shifting windows for representation, which allows for modeling at various scales and thus boosts the performance of vision transformers. In addition to image classification, transformers have been applied to other computer vision tasks, such as object detection \cite{carion2020end,zhu2020deformable}, semantic segmentation \cite{chen2021pre}, and video recognition \cite{neimark2021video}. 

Although these vision transformer models have great potential on computer vision tasks, their powerful representation capabilities are obtained based on the complicated model architectures, which makes them supremely challenging to deploy on resource-constrained devices and execute real-time inference \cite{tang2021patch,liu2021post}. Thus, model compression is a necessary and promising solution to facilitate their real-world applications.

\subsection{Model Quantization}
Model quantization, which reduces the memory footprint and computational overhead of the models by decreasing the representation precision of the weights and activations, is an effective approach to compressing models in a hardware-friendly manner \cite{krishnamoorthi2018quantizing,gholami2021survey}. The mainstream methods exploit quantization-aware training strategies to compensate the accuracy degradation caused by discretization \cite{zhang2018lq,li2019additive,Li2022IViTIQ,choi2018pact,esser2019learned}, and they use the straight-through estimator \cite{bengio2013estimating} to approximate the gradient back-propagation of the quantized model. However, these methods rely heavily on the original dataset for re-training/fine-tuning, rendering them inapplicable in many scenarios where the original data is not available \cite{cai2020zeroq,zhang2021diversifying}. In addition, several post-training quantization methods have been proposed to reduce the fine-tuning cost \cite{jacob2018quantization,li2019fully,choukroun2019low,wu2020easyquant,nagel2020up,li2021brecq}, including schemes for vision transformers \cite{liu2021post,yuan2021ptq4vit,lin2021fq}, but they still require a small amount of real data for calibration and cannot achieve complete data-free.

\textbf{Data-free quantization}, which compresses models without access to any real data, can potentially address the above issues, and thus has received increasing attention. ZeroQ \cite{cai2020zeroq} proposes BN regularization to generate samples based on the real-data statistics encoded in the BN layers of the pre-trained FP model, and then use them to fine-tune the model parameters. DSG \cite{zhang2021diversifying} presents an improved BN regularization scheme that utilizes slack distribution alignment and layerwise sample enhancement to address the homogenization of the generated samples. GDFQ \cite{xu2020generative} and IntraQ \cite{zhong2021intraq} introduce category label information to generate class-conditional samples, further pushing the limit of data-free quantization. However, these methods are only applicable to CNNs, because there is no key structure BN in vision transformers and the LN they employ does not contain any features of the original training data. As a result, there is now a gap in the data-free quantization community of vision transformers.

\section{Methodology}
In this section, we first introduce the computational process of vision transformers and the uniform quantization strategy in the preliminaries. Our insights and motivations of the proposed PSAQ-ViT are then presented, followed by a detailed introduction to the designed patch similarity metric for sample generation. Finally, the overall quantization pipeline for vision transformers is summarized and presented.

\subsection{Preliminaries}
A standard transformer's input is a one-dimensional sequence of token embeddings. For vision transformers, an image $I$ is reshaped into a sequence of flatted 2D patches, and each patch is then mapped to the hidden size $d$ by a linear projection to obtain the input vectors $X\in \mathbb{R}^{N\times d}$. Here, $N$ is the number of patches.

The vectors $X$ are then input into transformer layers, which are a stack of blocks composed of a multi-head self-attention (MSA) module and a multi-layer perceptron (MLP) module apiece. First, MSA calculates the attention between different patches to extract feature representations with global information as follows:
\begin{equation}
	\begin{aligned}
		\text{MSA}(X)&=\text{Concat}(\text{head}_1,\cdots,\text{head}_H)W^o \\
		\text{where head}_i &= \text{Attn}(Q_i,K_i,V_i)=\text{softmax}(\frac{Q_iK_i^T}{\sqrt{d}})V_i
	\end{aligned}
\end{equation}
where $H$ is the number of attention heads. Here, query $Q_i$, key $K_i$, and value $V_i$ are computed by linear projections using matrix multiplication, $i$.$e$., $Q_i=XW_i^Q$, $K_i=XW_i^K$, $V_i=XW_i^V$. Then, the output of MSA is fed into MLP, which contains two fully-connected layers for feature mapping and information fusion.

As we can see, in vision transformers, most computational costs are derived from the large matrix multiplication in MSA and MLP modules. Thus, we intend to quantize all the parameters in matrix multiplication, including both weights and activations. 
In this paper, we perform the uniform quantization strategy, which is the most popular and hardware-friendly method and is defined as follows:
\begin{equation}
	\theta^q = \lfloor\frac{\text{clip}(\theta^p,q_0,q_{2^k-1})-q_0}{\Delta}\rceil, \; \text{where} \, \Delta = \frac{q_{2^k-1}-q_0}{2^k-1}
	\label{eq:quant}
\end{equation}
where $\theta^p$ and $\theta^q$ denote the parameters of the FP model and the quantized model, respectively. Here, $q_0$ and $q_{2^k-1}$ are clipping values that determine the quantization scales, $\lfloor\cdot\rceil$ is the round operator, and $k$ is the quantization bit-precision.

\subsection{Our Insights}
As mentioned before, the main challenge of data-free quantization for vision transformers is that they do not have BN layers that store information about the original training data, resulting in no available absolute value prior information for sample generation and thus no efficient calibration of quantization parameters.
Therefore, our interest is to mine deeper into the prior information of the pre-trained vision transformer models and thus explore a reliable \emph{relative value} metric that can well describe the general difference between Gaussian noise and the real image, so that we can reduce this difference to make the Gaussian noise approximate the real image.

Since the self-attention module is the unique structure of vision transformers, its powerful feature extraction capability is believed to contain a certain amount of original data information. Hence, we provide an in-depth analysis of the training process of the self-attention module, and then we observe that the reason the model can make good decisions is that the self-attention module can distinguish the foreground from the background of the training data, thus allocating more attention to the foreground that is more important for the decision. Since the input of vision transformers are independent vectors mapped by 2D patches, the responses of the self-attention module to different patches are significantly different, $i$.$e$., the foreground patches receive more attention.

When the pre-trained model executes inference, real images consistently produce the above features, while Gaussian noise, whose foreground is not easily extracted, does not have a similar capability and inevitably leads to homogeneous responses, as shown in Fig. \ref{fig:1}. Note that the real images here are only used to verify the general difference ($e$.$g$., a certain metric of the real images is always larger than that of Gaussian noise), and they will not be involved in any subsequent process.
Therefore, this general difference can indirectly represent the prior information of vision transformers and thus can be used to design the relative value metric to guide the sample generation.

\subsection{Patch Similarity Metric}
Based on the above insights, we aim to design a reliable metric that can measure
the diversity of the self-attention module's responses. 
For the $l$-th layer in vision transformers, the output of the MSA module is defined as $O_l \in \mathbb{R}^{B\times H\times N\times d}$ ($l\in \{1,\cdots,L\}$), where each dimension denotes the batch size, number of heads, number of patches, and hidden size, respectively.
To simplify the expression, we ignore the batch dimension, $i$.$e$., $O_l \in \mathbb{R}^{H\times N\times d}$.

Due to the relative value metric, it is necessary to first normalize $O_l$ to ensure the fairness of the comparison. We accomplish this by calculating the cosine similarity between each subspace vector in the patch dimension, specifying the data range at [-1, 1], as follows:
\begin{equation}
	\Gamma_l(u_i,u_j) = \frac{u_i\cdot u_j}{||u_i||\;||u_j||}
\end{equation}
where the numerator represents the inner product of the vectors, and $||\cdot||$ denotes the $l_2$ norm. Here, $u_i,u_j \in \mathbb{R}^{H\times d}$ ($i,j\in \{1,\cdots, N\}$) is the $i$-th/$j$-th vector in the patch dimension of $O_l$, and $\Gamma_l(u_i,u_j)$ represents the cosine similarity between $u_i$ and $u_j$.
After pairwise calculations, we obtain the $l$-th layer's cosine similarity matrix $\bm{\Gamma}_l=[\Gamma_l(u_i,u_j)]_{N\times N}$, which is a symmetric matrix and is termed as patch similarity. The diversity of patch similarity can potentially represent the diversity of the original data, which not only elegantly achieves data normalization, but also has the additional advantage of achieving reasonable $\frac{Hd}{N}$-fold  dimensionality reduction ($\mathbb{R}^{H\times N\times d}\to \mathbb{R}^{N\times N}$). For instance, for the ViT-B model, the amount of data is reduced by a factor of 3.92, which can greatly improve the subsequent computational efficiency.

Then, the diversity of patch similarity is measured by the information entropy, which can represent the amount of information expressed by the data.
To ensure gradient back-propagation, we calculate the differential entropy that has a continuous nature as follows:
\begin{equation}
	H_l = -\int \hat{f_h}(x) \cdot \log \left[ \hat{f_h}(x) \right] dx
\end{equation}
where $\hat{f_h}(x)$ is the continuous probability density function of $\bm{\Gamma}_l$, which is obtained using kernel density estimation as follows:
\begin{equation}
	\hat{f_h}(x) = \frac{1}{M}\sum_{m=1}^{M}K_h(x-x_m)=\frac{1}{Mh}\sum_{m=1}^{M}K(\frac{x-x_m}{h})
\end{equation}
where $K(\cdot)$ is the kernel ($e$.$g$. normal kernel), $h$ is the bandwidth, $x_m$ ($m\in \{1,\cdots,M\}$) is a training point drawn from $\bm{\Gamma}_l$ and is the center of a kernel, and $x$ is the given test point.

Finally, we sum the differential entropy of each layer to account for the diversity of patch similarity across all layers, and since it is to be maximized, the Patch Similarity Entropy loss is defined as follows:
\begin{equation}
	\mathcal{L}_{PSE} = -\sum_{l=1}^{L}H_l
	\label{eq:l_pse}
\end{equation}

\subsection{The Overall Pipeline}
The whole process of PSAQ-ViT is performed in two stages: first, the Gaussian noise is optimized according to the loss function, which is designed based on the prior information of the pre-trained model, to generate ``realistic" samples; second, the generated samples are utilized to calibrate the quantization parameters, thus realizing the vision transformer quantization with no real data participation. These two stages are described in detail below.

\textbf{Sample generation:} In the sample generation stage, in addition to our proposed patch similarity entropy loss $\mathcal{L}_{PSE}$, which has the greatest contribution to the performance, the optimization objective for sample generation also contains two auxiliary image priors: one-hot loss $\mathcal{L}_{OH}$ and total variance loss $\mathcal{L}_{TV}$, which can ensure more stable convergence to effective images.

One-hot loss is a popular class prior that describes the class boundary information and motivates the generated images to be predicted to a pre-defined category $c$ \cite{xu2020generative,zhong2021intraq}. Specifically, it encourages to minimize the cross entropy loss as follows:
\begin{equation}
	\mathcal{L}_{OH} = CE(P(I),c)
	\label{eq:l_oh}
\end{equation}
where $P(I)$ is the predicted result of the pre-trained model for image $I$.

Total variance loss is a pixel-level smoothing regularization term for images and can further improve the image quality \cite{yin2020dreaming}, which is defined as follows:
\begin{equation}
	\mathcal{L}_{TV} = \iint |\nabla I(\tau_1,\tau_2)|d\tau_1d\tau_2
	\label{eq:l_tv}
\end{equation}
where $\nabla I(\tau_1,\tau_2)$ denotes the gradient of the image $I$ at $(\tau_1,\tau_2)$.

We combine the above three loss functions to obtain the final objective function for sample generation as follows:
\begin{equation}
	\begin{aligned}
			\mathcal{L}_{G} & = \mathcal{L}_{PSE} + \alpha \mathcal{L}_{prior} \\
			& = \mathcal{L}_{PSE} + \alpha_1 \mathcal{L}_{OH} + \alpha_2 \mathcal{L}_{TV}
	\end{aligned}
\label{eq:l_g}
\end{equation}
where $\alpha_1$ and $\alpha_1$ are the balance coefficients.

\textbf{Quantization parameter calibration:} In the parameter calibration stage, the weight parameters are fixed and can be calibrated directly, thus the generated samples are only utilized to determine the clipping values ($q_0$ and $q_{2^k-1}$) for the activations of each layer to get rid of outliers and to better represent the majority of the given parameters. Note that the calibration process is performed in the form of post-training quantization and does not require resource-consuming fine-tuning.
The overall pipeline is summarized in Algorithm \ref{alg:PSAQ}.
\begin{algorithm}[t]
	\caption{The PSAQ-ViT Pipeline}
	\label{alg:PSAQ}
	\SetAlgoLined
	\KwIn{A pre-trained FP vision transformer $P$ with parameters $\theta^p$.}
	
	\KwOut{A quantized vision transformer $Q$ with parameters $\theta^q$.}
	
	Initialize the quantized model $Q$ by Eq. (\ref{eq:quant});
	
	Randomly produce Gaussian noise $I_G\thicksim \mathcal{N}(0,1)$;
	
	\textbf{\textcolor{teal}{\# Stage 1: Sample generation}}
	
	\For{$t=1,2,\ldots$}{
		
		Input $I_G$ into the pre-trained FP model $P$;
		
		Calculate $\mathcal{L}_{PSE}$ by Eq. (\ref{eq:l_pse});
		
		Calculate $\mathcal{L}_{OH}$ and $\mathcal{L}_{TV}$ by Eq. (\ref{eq:l_oh}) and Eq. (\ref{eq:l_tv});
		
		Combine three losses to obtain $\mathcal{L}_{G}$ by Eq. (\ref{eq:l_g});
		
		Update $I_G$ by back-propagation of $\mathcal{L}_{G}$;
	}

	\textbf{\textcolor{teal}{\# Stage 2: Quantization parameter calibration}}
	
	Get the generated samples $I = I_G$;
	
	Input $I$ into the quantized model $Q$;
	
	Determine the clipping values of the activations in $Q$;
	
\end{algorithm}

\section{Experiments}
In this section, PSAQ-ViT is evaluated on various benchmark models for the large-scale image classification task. To the best of our knowledge, there is no published work on data-free quantization of vision transformers, thus the effectiveness of our method is demonstrated by comparing the quantized model calibrated with real images and Gaussian noise at the same settings. Furthermore, ablation studies are conducted to verify the validity of the proposed patch similarity entropy loss.

\subsection{Implementation Details}
\textbf{Models and Datasets:} We evaluate PSAQ-ViT on various popular vision transformer models, including ViT \cite{dosovitskiy2020image}, DeiT \cite{touvron2021training}, and Swin \cite{liu2021swin}. The dataset we adopt is ImageNet (ILSVRC-2012) \cite{krizhevsky2012imagenet} for the large-scale image classification task which contains 1000 categories of images (224$\times$224 pixels). The pre-trained models are all obtained from timm\footnote[1]{https://github.com/rwightman/pytorch-image-models}.

\textbf{Experimental settings:} All implementations of PSAQ-ViT are done on PyTorch. To demonstrate the validity of our generated images, we employ the most basic quantization parameter calibration method. For weights, symmetric uniform quantization is applied, and the calibration strategy is fixed to Vanilla MinMax; for activations, asymmetric uniform quantization is applied, and the default calibration strategy is Vanilla MinMax if not specifically declared. In all our experiments, the number of images used for calibration is 32. $\alpha_1$ and $\alpha_2$ are set to 1.0 and 0.05 after a simple grid search, respectively, and their selection had little effect on the final performance.

\subsection{Analysis of generated samples}
\begin{figure}[t]
	\centering
	\includegraphics[width=0.57\linewidth]{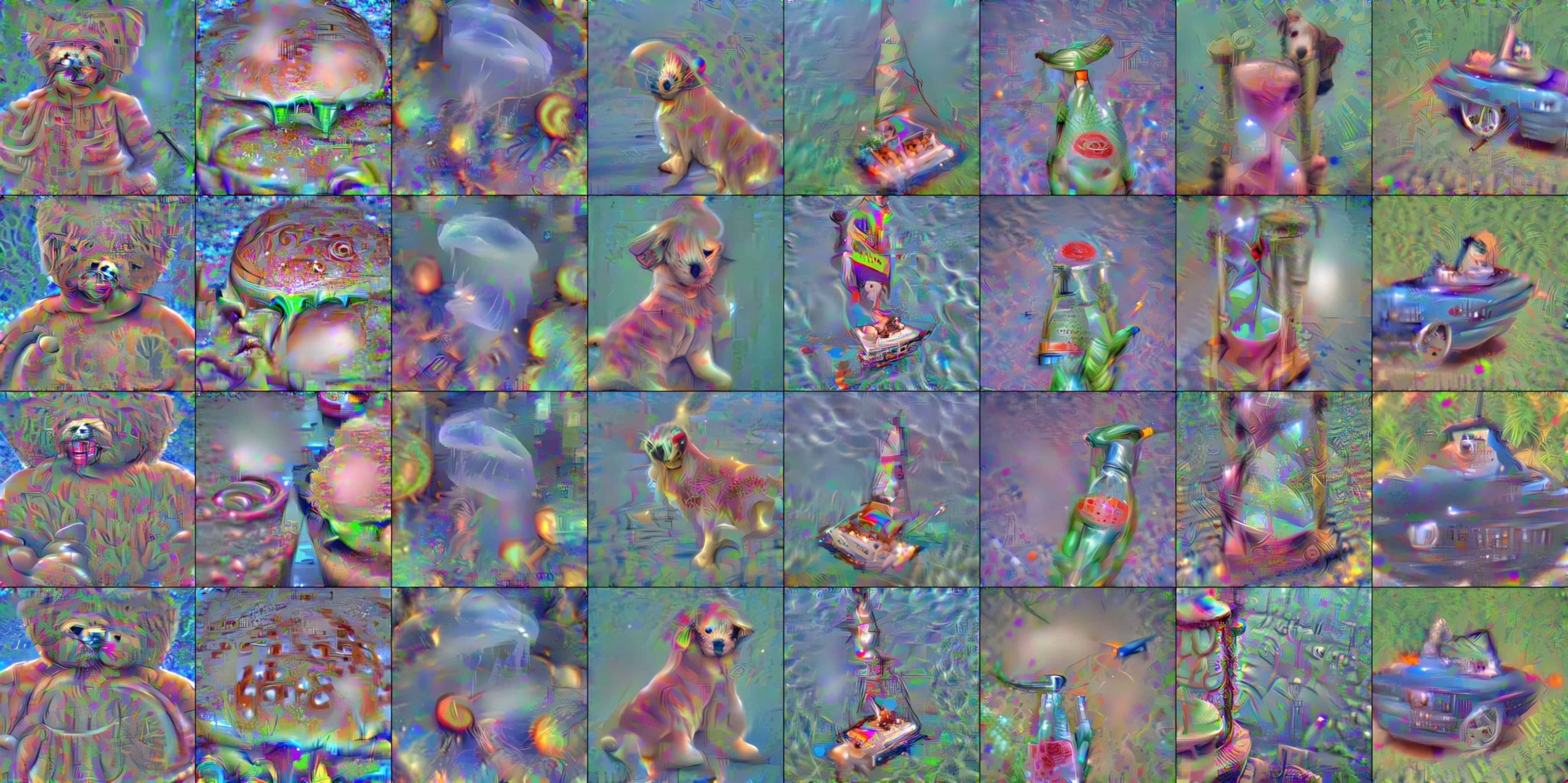}
	\caption{Class-conditional samples (224$\times$224 pixels) generated by PSAQ-ViT, given only a pre-trained ViT-B model on ImageNet and no additional information.}
	\label{fig:generated_img}
\end{figure}

Fig. \ref{fig:generated_img} shows the visualization results of the generated images (224$\times$224 pixels), which are obtained based on the ViT-B model pre-trained on ImageNet dataset. Since we use the class prior $\mathcal{L}_{OH}$ in the image generation process, we present them by category, and different images in a category are produced by using different random seeds when initializing the Gaussian noise. It should be highlighted that these images require only a pre-trained model, and not any additional information, especially the original data or any absolute value metrics. Thanks to the proposed optimization objective $\mathcal{L}_{PSE}$, the generated ``realistic" images can clearly distinguish the foreground from the background, and the foreground is rich in semantic information.
Moreover, according to the subsequent quantization experiments, this excellent property of easily extracting the foreground will have a positive feedback effect on the calibration of the quantization parameters, making the generated images achieve better performance than the real images.
\begin{figure}[t]
	\centering
	\includegraphics[width=0.8\linewidth]{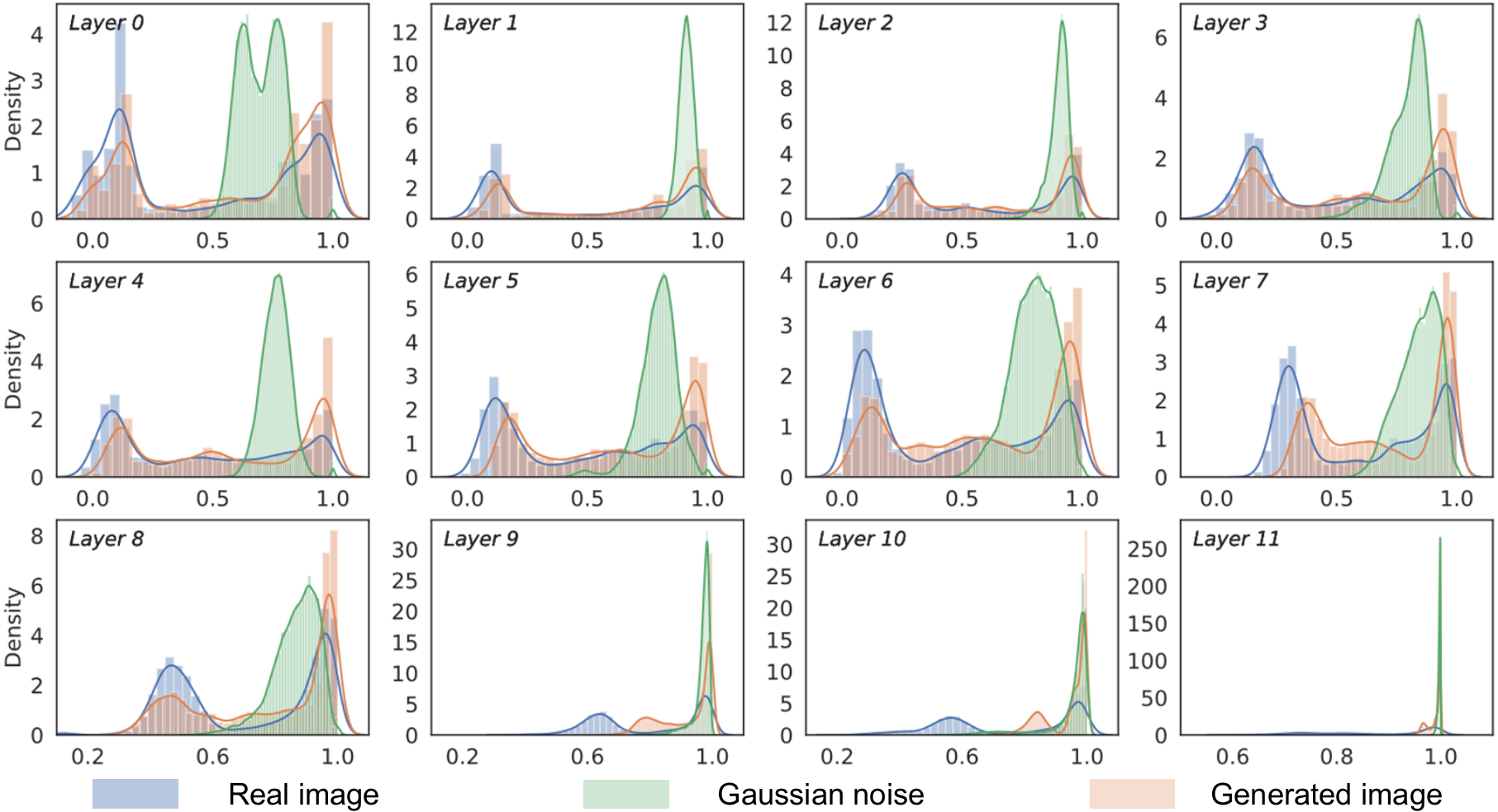}
	\caption{Comparison of the kernel density curves of the patch similarity for each layer in ViT-B model when the input is the real image, Gaussian noise, and the generated image. The x-axis represents the values of patch similarity. As we can see, the density of each layer corresponding to Gaussian noise shows a concentrated unimodal shape, while the generated image and the real image have similar properties, producing the density with a dispersed bimodal shape.}
	\label{fig:compare12}
\end{figure}

In addition, since we consider the patch similarity entropy of all layers in Eq. (\ref{eq:l_pse}), the comparison experiments of the kernel density curves of the patch similarity for each layer in ViT-B model when the input is the real image, Gaussian noise, and the generated image are conducted, as shown in Fig. \ref{fig:compare12}. For the responses to Gaussian noise, the kernel density curves all show a concentrated unimodal shape and the central value of the curve is high, indicating a high degree of similarity between each patch of Gaussian noise and thus a full classification as background or foreground.
Fortunately, the kernel density curves corresponding to our generated images are very approximate to those corresponding to the real images. They all show a dispersed bimodal shape, indicating a high diversity of responses, and the left and right peaks of curves describe inter- and intra-category similarity, respectively, which is in line with the expectation that the images can easily be distinguished between foreground and background.

\subsection{Quantization results}
Here, we employ the proposed PSAQ-ViT to quantify the ViT-S, ViT-B, DeiT-T, DeiT-S, DeiT-B, Swin-T, and Swin-S models on large-scale ImageNet dataset, and the results are reported in Table \ref{exp:ImageNet}. Since, to the best of our knowledge, there is no published work on data-free quantization of vision transformers, we set a reasonable baseline for our experiments on our own. \emph{Standard} and \emph{Gaussian noise} denote using real images and Gaussian noise to calibrate the quantization parameters, respectively. Note that all experiments differ only in the images used to calibrate the quantization parameters, and all other settings are the same, such as the calibration strategy and the number of images. Moreover, to demonstrate the robustness of the method, we evaluate different quantization precisions, including W4/A8 and W8/A8.

First, it should be emphasized that PSAQ-ViT can achieve \textbf{better performance} than Standard, which requires the real data, on all the aforementioned models, indicating that the generated images are even more effective than the real ones for parameter calibration. The main reason is that the sample generation is based on the prior information in the self-attention module, $i$.$e$., facilitating the distinction between foreground and background in images, and then when these samples are utilized to calibrate the quantization parameters, they in turn reinforce the functionality of the self-attention module, thus acting as positive feedback that can reduce the activation outliers to some extent and therefore improve the tolerance to parameter clipping.

The quantization results of each model are discussed in detail below. We begin by discussing the quantization results of ViT-S and ViT-B models. Because we use the vanilla quantization strategy and these models are very sensitive to quantization, different methods can all lead to noticeable accuracy degradation. Despite this, our method achieves the best performance at the same settings, $e$.$g$. for quantization of ViT-S, our method improves by 0.93\% and 1.17\% over Standard at W4/A8 and W8/A8 settings, respectively.
\setlength{\tabcolsep}{5pt}
\begin{table}[t]\scriptsize
	\begin{center}
		\caption{Quantization results on ImageNet dataset. Standard, Gaussian noise, and PSAQ-ViT calibrate the quantization parameters with real images, Gaussian noise, and generated images, respectively, and all other settings are the same. As we can see, PSAQ-ViT can even be more superior than real-data-driven Standard. Here, ``No Data" indicates that no real data participate in the quantization process, ``Prec." denotes the quantization precision, ``W$x$/A$y$" denotes quantifying the weights and activations to $x$-bit and $y$-bit, respectively, and ``Top-1" is the Top-1 test accuracy of the quantized vision transformer.}
		\label{exp:ImageNet}
		\begin{tabular}{c|cc|cc|cc}
			\toprule
			Model & Method & No Data & Prec. & Top-1(\%) & Prec. & Top-1(\%) \\
			
			\midrule
			
			\multirow{3}*{\tabincell{c}{ViT-S \\ (81.39)}} 
			& Standard & $\times$ & W4/A8 & 19.91 & W8/A8 & 30.28 \\
			& Gaussian noise & $\checkmark$ & W4/A8 & 15.60 & W8/A8 & 25.22 \\
			& PSAQ-ViT(ours) & $\checkmark$ & W4/A8 & \textbf{20.84} & W8/A8 & \textbf{31.45} \\
			
			\cmidrule{1-7}	
			\multirow{3}*{\tabincell{c}{ViT-B \\ (84.53)}} 
			& Standard & $\times$ & W4/A8 & 24.76 & W8/A8 & 36.65 \\
			& Gaussian noise& $\checkmark$ & W4/A8 & 19.45 & W8/A8 & 31.63 \\
			& PSAQ-ViT(ours) & $\checkmark$ & W4/A8 & \textbf{25.34} & W8/A8 & \textbf{37.36} \\

			\cmidrule{1-7}
			\multirow{3}*{\tabincell{c}{DeiT-T \\ (72.21)}} 
			& Standard & $\times$ & W4/A8 & 65.20 & W8/A8 & 71.27 \\
			& Gaussian noise & $\checkmark$ & W4/A8 & 7.80 & W8/A8 & 10.55\\
			& PSAQ-ViT(ours) & $\checkmark$ & W4/A8 & \textbf{65.57} & W8/A8 & \textbf{71.56}\\

			\cmidrule{1-7}
			\multirow{3}*{\tabincell{c}{DeiT-S \\ (79.85)}} 
			& Standard & $\times$ & W4/A8 & 72.10 & W8/A8 & 76.00 \\
			& Gaussian noise & $\checkmark$ & W4/A8 & 13.30 & W8/A8 & 18.16 \\
			& PSAQ-ViT(ours) & $\checkmark$ & W4/A8 & \textbf{73.23} & W8/A8 & \textbf{76.92} \\
			
			\cmidrule{1-7}
			\multirow{3}*{\tabincell{c}{DeiT-B \\ (81.85)}} 
			& Standard & $\times$ & W4/A8 & 76.25 & W8/A8 & 78.61 \\
			& Gaussian noise & $\checkmark$ & W4/A8 & 11.09 & W8/A8 & 14.72 \\
			& PSAQ-ViT(ours) & $\checkmark$ & W4/A8 & \textbf{77.05} & W8/A8 & \textbf{79.10} \\
			
			\cmidrule{1-7}
			\multirow{3}*{\tabincell{c}{Swin-T \\ (81.35)}} 
			& Standard & $\times$ & W4/A8 & 70.16 & W8/A8 & 74.22 \\
			& Gaussian noise & $\checkmark$ & W4/A8 & 0.52 & W8/A8 & 0.62 \\
			& PSAQ-ViT(ours) & $\checkmark$ & W4/A8 & \textbf{71.79} & W8/A8 & \textbf{75.35} \\
			
			\cmidrule{1-7}
			\multirow{3}*{\tabincell{c}{Swin-S \\ (83.20)}} 
			& Standard & $\times$ & W4/A8 & 73.33 & W8/A8 & 75.19 \\
			& Gaussian noise & $\checkmark$ & W4/A8 & 5.43 & W8/A8 & 5.66 \\
			& PSAQ-ViT(ours) & $\checkmark$ & W4/A8 & \textbf{75.14} & W8/A8 & \textbf{76.64} \\

			\bottomrule
		\end{tabular}
	\end{center}
\end{table}
\setlength{\tabcolsep}{1.4pt}

DeiT has the same model structure as ViT but with a different training strategy; however, the quantization perturbation on the performance of DeiT is significantly smaller compared to ViT. When the calibration image is Gaussian noise, the representation capability of the quantization model decreases sharply, and its prediction accuracy decreases severely. In comparison, our proposed PSAQ-ViT can achieve very excellent performance. For the W8/A8 quantization of DeiT-T, our method achieves 4-fold compression with almost lossless accuracy (only 0.65\% accuracy degradation).
PSAQ-ViT is 1.13\% and 0.92\% higher than real-data-driven Standard in the quantization of W4/A8 and W8/A8 for DeiT-S, respectively. The results of the quantization of DeiT-B, which are similar to those of the previous models, show that our method also achieves the best performance, with an improvement of 0.8\% and 0.49\% over Standard at W8/A8 and W4/A8 settings, respectively.

The proposed PSAQ-ViT still maintains a high level of robustness to the Swin models. When the quantized Swin-T is calibrated with Gaussian noise, the model performance becomes almost infeasible; nevertheless, our method can guarantee a small performance degradation. In addition, PSAQ-ViT is very quantization-friendly for Swin-S, achieving substantial performance improvements over Standard, with gains of 1.81\% at W4/A8 and 1.45\% at W8/A8, respectively.

\subsection{Results of combining with post-training quantization}
To demonstrate the generality of the proposed method, we evaluate the results of combining PSAQ-ViT with post-training quantization methods, which further improves the performance of quantization. Specifically, instead of using vanilla MinMax, we use EMA \cite{jacob2018quantization}, Percentile \cite{li2019fully}, and OMSE \cite{choukroun2019low} to determine the clipping values for activations. Among them, EMA employs a moving average mechanism to smooth the maximum and minimum values of the tensors; Percentile clips the tensors according to the percentile of the parameters (1e-5 percentile is used in the experiments); OMSE minimizes the quantization error to determine the tensors' clipping values. In addition, the experimental results are compared with PTQ-ViT \cite{liu2021post}, which is the state-of-the-art (SOTA) ranking-aware post-training quantization method for vision transformers. Note that PTQ-ViT has higher computational complexity and requires the assistance of 1000 real images, while our method requires only 32 generated images.

\setlength{\tabcolsep}{5pt}
\begin{table}[t]\scriptsize
	\begin{center}
		\caption{Quantization results of combining with post-training quantization methods on ImageNet dataset. Our PSAQ-ViT combined with simple post-training quantization methods, including EMA \cite{jacob2018quantization}, Percentile \cite{li2019fully}, and OMSE \cite{choukroun2019low}, can achieve comparable performance to the SOTA ranking-aware post-training method \cite{liu2021post} that has high computational complexity and requires the assistance of 1000 real images.}
		\label{exp:PTQ}
		\begin{tabular}{c|ccc|cc}
			\toprule
			Model & Method & Strategy & No Data &  Prec. & Top-1(\%) \\
			
			\midrule
			
			\multirow{6}*{\tabincell{c}{DeiT-S \\ (79.85)}} 
			& PTQ-ViT & Ranking-Aware & $\times$ & W8/A8 & 77.47 \\
			\cmidrule{2-6}
			&\multirow{4}*{PSAQ-ViT(ours)} 
			& Vanilla & $\checkmark$ &  W8/A8 & 76.92 \\
			\cmidrule{3-6}
			& & EMA & $\checkmark$ & W8/A8 & 77.12 \\
			& & Percentile & $\checkmark$ & W8/A8 & \textbf{77.31} \\
			& & OMSE & $\checkmark$ & W8/A8 & 76.94 \\
			
			\cmidrule{1-6}	
			\multirow{6}*{\tabincell{c}{DeiT-B \\ (81.85)}} 
			& PTQ-ViT & Ranking-Aware & $\times$ & W8/A8 & 80.48 \\
			\cmidrule{2-6}
			&\multirow{4}*{PSAQ-ViT(ours)} 
			& Vanilla & $\checkmark$ & W8/A8 & 79.10 \\
			\cmidrule{3-6}
			& & EMA & $\checkmark$ & W8/A8 & 79.99 \\
			& & Percentile & $\checkmark$ & W8/A8 & 79.42 \\
			& & OMSE & $\checkmark$ & W8/A8 & \textbf{80.26} \\

			\bottomrule
		\end{tabular}
	\end{center}
\end{table}
\setlength{\tabcolsep}{1.4pt}

The quantization results are reported in Table \ref{exp:PTQ}. PSAQ-ViT, when combined with post-training quantization methods, can achieve comparable performance to PTQ-ViT. Meanwhile, different models have different preferences for different calibration strategies. For instance, PSAQ-ViT combined with Percentile shows the best performance on DeiT-S, while DeiT-B achieves the highest accuracy when using OMSE to calibrate the parameters.

\subsection{Ablation Studies}

We perform ablation studies on DeiT-S and DeiT-B models to demonstrate the effectiveness of different loss functions used for sample generation, and the results are shown in Table \ref{exp:ablation}. We first analyze the experimental results of DeiT-S. Not using any loss function, $i$.$e$., calibrating directly with Gaussian noise, certainly leads to an unexpected decrease in accuracy; when only $\mathcal{L}_{OH}$ and $\mathcal{L}_{TV}$ are used to optimize the noise, the accuracy of the quantized model is still far from satisfactory. Using only the proposed patch similarity entropy loss $\mathcal{L}_{PSE}$ can guarantee good quantization performance, and since it is completely decoupled from the other losses, it can be easily combined with them to achieve better results where it has the largest contribution to the final performance.
A similar analysis also applies to DeiT-B. It is well demonstrated that our designed $\mathcal{L}_{PSE}$ has an essential driving effect on the quality improvement of the generated images, thus ensuring an effective calibration of the quantization parameters.

\setlength{\tabcolsep}{8pt}
\begin{table}[t]\scriptsize
	\begin{center}
		\caption{Ablation study of different loss functions for sample generation. $\mathcal{L}_{PSE}$ has the largest contribution to the final results. In addition, it is fully decoupled from other losses, thus it can further improve performance in combination with other losses.}
		\label{exp:ablation}
		\begin{tabular}{c|ccc|cc}
			\toprule
			Model & $\mathcal{L}_{PSE}$ & $\mathcal{L}_{OH}$ & $\mathcal{L}_{TV}$ &  Prec. & Top-1(\%) \\
			
			\midrule
			
			\multirow{6.5}*{\tabincell{c}{DeiT-S \\ (79.85)}} 
			& $\times$ & $\times$ & $\times$ & W8/A8 &  18.16 \\
			& $\times$ & $\checkmark$ & $\checkmark$ & W8/A8 &  65.66 \\
			\cmidrule{2-6}
			& $\checkmark$ & $\times$ & $\times$ & W8/A8 &  74.07 \\
			& $\checkmark$ & $\checkmark$ & $\times$ & W8/A8 &  75.39 \\
			& $\checkmark$ & $\times$ & $\checkmark$ & W8/A8 &  75.28 \\
			& $\checkmark$ & $\checkmark$ & $\checkmark$ & W8/A8 &  \textbf{76.92} \\

			\cmidrule{1-6}	
			\multirow{6.5}*{\tabincell{c}{DeiT-B \\ (81.85)}} 
			& $\times$ & $\times$ & $\times$ & W8/A8 &  14.72 \\
			& $\times$ & $\checkmark$ & $\checkmark$ & W8/A8 &  67.95 \\
			\cmidrule{2-6}
			& $\checkmark$ & $\times$ & $\times$ & W8/A8 & 78.07  \\
			& $\checkmark$ & $\checkmark$ & $\times$ & W8/A8 &  78.50 \\
			& $\checkmark$ & $\times$ & $\checkmark$ & W8/A8 &  78.61 \\
			& $\checkmark$ & $\checkmark$ & $\checkmark$ & W8/A8 &  \textbf{79.10} \\

			\bottomrule
		\end{tabular}
	\end{center}
\end{table}
\setlength{\tabcolsep}{1.4pt}

We also perform efficiency analysis of the two stage of PSAQ-ViT divided in Algorithm \ref{alg:PSAQ}, as shown in table \ref{exp:time}. The whole process takes less than 4 min on an RTX 3090 GPU and most time is spent in the image generation, since the parameter calibration without training produces small overhead.
\setlength{\tabcolsep}{5pt}
\begin{table}[t]\scriptsize
	\caption{Efficiency analysis of PSAQ-ViT on DeiT-B, which spends less than 4 min on an RTX 3090 GPU, with the majority spent on image generation.}
	\label{exp:time}
	\begin{center}
		\begin{tabular}{c|c|cc|cc}
			\toprule
			Model &  Image Generation(s) & \multicolumn{4}{c}{Quantization Calibration(s)}\\
			
			\midrule
            DeiT-B & 227 & Vanilla &  0.17 & OMSE & 0.41 \\
			
			\bottomrule
		\end{tabular}
	\end{center}
\end{table}
\setlength{\tabcolsep}{1.4pt}

\section{Conclusions}

In this paper, we propose PSAQ-ViT, a Patch Similarity Aware data-free Quantization framework for Vision Transformers. First, we perform an in-depth analysis of the unique properties of the self-attention module, revealing a general difference in its processing of Gaussian noise and real images. Based on this insight, we design a relative value metric to optimize the Gaussian noise to approximate the real image. Specifically, we use the differential entropy of patch similarity calculated via kernel density estimation to represent the diversity of the self-attention module's responses, then maximize the entropy to optimize the Gaussian noise, and finally utilize the generated ``realistic" samples to efficiently calibrate the quantization parameters.
Extensive experiments and ablation studies are conducted to demonstrate the effectiveness of PSAQ-ViT and the proposed patch similarity entropy loss.
Thanks to the positive feedback effect of the generated images that are easily distinguished between foreground and background as analyzed in our paper, PSAQ-ViT can even outperform the real-data-driven methods at the same settings.

\section*{Acknowledgements}
This work was supported in part by the National Natural Science Foundation of China under Grant 62276255; in part by the Scientific Instrument Developing Project of the Chinese Academy of Sciences under Grant YJKYYQ20200045.

%
%
\bibliographystyle{splncs04}
\bibliography{egbib}
\end{document}